\pdfoutput=1
\documentclass[11pt]{article}
\usepackage{rotating} % <-- HERE

\usepackage[table,xcdraw]{xcolor}

\usepackage{emnlp2021}

\usepackage{float}

\usepackage{times}
\usepackage{latexsym}
\usepackage{url}
\usepackage{enumitem}
\usepackage{tablefootnote}
\usepackage{rotating}

\usepackage[T1]{fontenc}
\usepackage{booktabs}
\usepackage[utf8]{inputenc}

\usepackage{microtype}

\makeatletter
 \twocolumn 

\title{Open foundation models for Azerbaijani language}

\author{Jafar Isbarov* \\
  The George Washington University \\
  Department of Computer Science \\
  0000-0001-8404-2192 \\
  jafar.isbarov@gwmail.gwu.edu \\\And
  Kavsar Huseynova* \\
  Baku Higher Oil School \\
  Information Technology Department \\
  0009-0007-0362-9591 \\
  kavsar.huseynova.std@bhos.edu.az \\\AND
  Elvin Mammadov  \\
  Baku Higher Oil School \\
  Information Technology Department \\
  0009-0005-9237-9736 \\
elvin.mammadov.std@bhos.edu.az \\\And
  Mammad Hajili  \\
  Microsoft \\
  0000-0002-9522-2137 \\
mammadhajili@microsoft.com \\\AND
  Duygu Ataman \\
  New York University \\
  Department of Computer Science \\
  ataman@nyu.edu \\
  }

\begin{document}
\maketitle
% UNCOMMENT IF NOT ANONYMOUS
\begingroup\def\thefootnote{*}\footnotetext{Equal contribution}\endgroup

\begin{abstract}
The emergence of multilingual large language models has enabled the development of language understanding and generation systems in Azerbaijani. However, most of the production-grade systems rely on cloud solutions, such as GPT-4. While there have been several attempts to develop open foundation models for Azerbaijani, these works have not found their way into common use due to a lack of systemic benchmarking. This paper encompasses several lines of work that promote open-source foundation models for Azerbaijani. We introduce (1) a large text corpus for Azerbaijani, (2) a family of encoder-only language models trained on this dataset, (3) labeled datasets for evaluating these models, and (4) extensive evaluation that covers all major open-source models with Azerbaijani support.
\end{abstract}

\section{Introduction}
Large language models (LLMs) have seen a sudden rise in popularity in recent years. Both open-source and proprietary models have seen wide adoption across various industries. This boost has not been shared equally across different regions, however, mostly due to the slow osmosis of these technologies into low-resource languages. Azerbaijani language falls on the "other" side of this barrier, with its 24 million speakers worldwide.

While some models have a limited understanding of the Azerbaijani language, only paid models offered by OpenAI have seen some level of adoption in the industry. Open-source models are being created with multilingual or Azerbaijani-only capabilities, but the community is not as keen to adopt them. This is possibly due to the limited exploration of these models' potential. This paper encompassed several lines of work that share a common goal - promoting open-source foundational models for Azerbaijani. Our contributions are as follows:
\begin{enumerate}
    \item DOLLMA: A new text corpus of 651.1 million words in Azerbaijani that can be used for pre-training LLMs.
    \item aLLMA: A new family of BERT-class models trained on this dataset from scratch.
    \item Three labeled datasets that can be used for benchmarking foundation models in Azerbaijani:
    \begin{enumerate}[label*=\arabic*.]
        \item AZE-SCI: A text classification dataset.
        \item AZE-NSP: A next-sentence prediction dataset.
        \item CB-MCQ: A closed-book question-answering dataset.
    \end{enumerate}
    \item A benchmark for several natural language understanding (NLU) tasks in Azerbaijani. It contains our newly introduced models and other existing open-source alternatives.
\end{enumerate}

\subsection{Foundation Models}
While language modeling has a long history, transformer-based large foundation models can be considered a recent phenomenon. These models have a disproportionately high number of trainable parameters, made possible due to the highly parallelizable nature of the transformer architecture. Their development takes place in two stages: Pre-training and fine-tuning. Pre-training is performed on Web-scale text corpora, while fine-tuning is performed on smaller and higher-quality data to adapt the model to a specific task. \citep{Minaee2024LargeLM}

Foundation models exist for various modalities, including language, vision, and speech. Language foundation models are usually classified as encoder, decoder, or encoder-decoder models. Encoder models are used for tasks that require language understanding, such as sentiment analysis and extractive question-answering. Encoder-decoder and decoder-only models are better suited for generative tasks, such as machine translation and text summarisation. \textit{Our work concentrates on encoder-only models.} Our main inspiration is the BERT model family by \citep{devlin-etal-2019-bert} and its derivatives.

In the rest of the paper, a foundation model refers to a language model trained on a vast amount of unlabeled text data that can be fine-tuned for various downstream tasks. A large language model refers to a foundation language model with at least tens of millions of parameters.

\subsection{Modeling Azerbaijani}
The majority of LLMs are either monolingual English models or multilingual models that do not support Azerbaijani. Very few multilingual models support Azerbaijani, and only recently monolingual Azerbaijani models are beginning to emerge.

This slow progress can be explained by several factors. A smaller market and less investment is an obvious explanation, but the field faces more fundamental challenges that would not be immediately solved by more funding. One of these is the state of digitalization of the language. Most of the electronic books in Azerbaijani are scanned books. Only books published since the 1990s are written in the last version of the Azerbaijani Latin alphabet \footnote{There was an older version of the Azerbaijani Latin alphabet introduced by the Soviets in 1922. This followed several variations until 1939 when the alphabet was replaced with a Cyrillic alternative. Azerbaijan started the transition to an updated Latin alphabet in 1991, which was completed in 2001.}, which creates another barrier. Yet another challenge is the small size of the community that's devoted to the development of open-source language models for Azerbaijani. The challenges regarding digitalization and script differences are further discussed in the third section.

An idea that is often heard regarding Azerbaijani LLMs is that we can simply go for the models developed for Turkish since languages are so similar. Azerbaijani and Turkish languages are not as similar as it is publicly perceived. According to \citep{azturk}, Azerbaijanis scored 56\% of receptive intelligibility in spoken Turkish. Differences in written language are not any smaller. Based on the methodology offered by \citep{langsimilarity}, a 44\% similarity score has been calculated between the vocabularies of the two languages \footnote{\url{https://www.ezglot.com/most-similar-languages?l=aze}}. Due to these significant differences, Turkish LLMs are not useful in machine learning tasks for Azerbaijani.

The paper is structured as follows. The next section gives a brief overview of previous works on foundational language models, and language modeling on Azerbaijani. The third section introduces DOLLMA, a new text corpus, and outlines the methodology, challenges we faced, and future works. The fourth section introduces aLLMA, a new family of monolingual encoder-only language models. The fifth section introduces several benchmarks for evaluating encoder-only Azerbaijani language models. These benchmarks are used to evaluate newly introduced models, as well as existing alternatives. The sixth section presents these benchmarks' results.

\section{Previous works}

The use of neural networks for language modeling can be traced back to the early 2000s. \citep{NIPS2000_728f206c} and \citep{mikolov10_interspeech} had created neural networks that outperformed traditional state-of-the-art model. \citep{schwenk-etal-2006-continuous} uses neural networks for machine translation. 

These models and their derivatives were task-specific. The idea of creating a foundational language model that could later be adapted (i.e., fine-tuned) to specific tasks was popularized only after the introduction of the transformer architecture by \citep{transformers}. The earliest foundational language model that gained wide adoption was BERT by \citep{devlin-etal-2019-bert} and later variations like RoBERTa \citep{liu2019roberta}.

BERT was an encoder-only model, therefore more suitable for problems that could be formulated as a subset of the classification problem. Generative foundation models came out around the same time, in the example of 
GPT-1 \citep{gpt1}, GPT-2 \citep{gpt2}, and T5 \citep{t5}. While the GPT series continued with closed-source, enterprise models, other alternatives quickly emerged with superior performance. The most famous of these was the LLaMA series, which directly or indirectly resulted in the development of hundreds of open-source language models.  \citep{touvron2023llama}.

Early foundation models were trained on English text, but multilingual models quickly emerged. Google had released multilingual BERT alternatives, and mGPT by \citep{mgpt} was an early variation of the GPT architecture for multiple languages. XLM-RoBERTa by \citep{xlm-roberta} was a larger and more successful alternative to mGPT and was quickly adopted worldwide. 

XLM-RoBERTa was also one of the first (if not the first) foundation models that supported Azerbaijani. We are aware of only one academic work that has concentrated on the development of foundational language models for Azerbaijani. \citep{azroberta} have trained a RoBERTa model on the Azerbaijani split of the OSCAR dataset \citep{ortiz-suarez-etal-2020-monolingual}. This work is a first of its kind for Azerbaijani and a very valuable starting point. However, it does not concentrate on the development of a foundation model. Its main focus is improving model performance by text augmentation. Therefore, they do not perform a systematic evaluation of the model. They have released one RoBERTa model, without different sizes, which is yet another limiting factor in the adoption of the work. Unfortunately, this model has not been included in our evaluation benchmarks because they have not released a tokenizer that is compatible with their model.

There have also been some community attempts to create such open-source models. A series of RoBERTa models were developed by continuing the pre-training phase on a small Azerbaijani dataset \citep{mammad_hajili_roberta}. Alas Development Center has developed a series of decoder-only LLMs for Azerbaijani \footnote{\url{https://github.com/interneuron-ai/project-barbarossa}}, but they offer no explanation regarding their approach, and the models failed to pass initial sanity checks.

\section{Text corpus}
A large text corpus is a prerequisite for training a large language model. For reference, GPT-2 and RoBERTa both were trained on OpenWebText \citep{liu2019roberta}, consisting of 13.5 billion tokens, which is roughly equivalent to 10 billion words. Original BERT models were trained on 3.3. billion words. While these numbers have exploded in recent years, the success of these models suggests that similarly effective models can be trained on similarly sized datasets.

The largest corpora that existed at the beginning of our work were OSCAR, which contained 316 million words in Azerbaijani, and Colossal Clean Crawled Corpus (C4) with 1.7 billion words. Introduced by \citep{c4},  C4 is one of the most widely used datasets in the pretraining stage of LLMs. C4 is labeled by language and contains 1.83 million documents tagged as Azerbaijani. Upon further inspection, however, we discovered a significant portion of this text is not only in different languages, but also in different alphabets (Armenian, Georgian, and Cyrillic). In addition, the C4 dataset contains a significant amount of informal text. This can be a valuable resource, but it is outside the scope of our work. Considering all of these points, we decided against using it. OSCAR \citep{ortiz-suarez-etal-2020-monolingual} dataset is also derived from CommonCrawl. It suffers from the same problems, so it was not included in our corpus either.

\begin{table*}
\centering
\begin{tabular}{lllll}
\toprule
Data source & Word count & Upscale & Final count & Source \\ \midrule
English Wikipedia & 194.0M & 4 & 776.0M & \citep{bhos_2024} \\
Azerbaijani Wikipedia & 40.0M & 6 & 245.0M & \citep{allma_lab_2024_azwiki} \\
News & 238.9M & 1 & 238.9M & BHOS AI R\&D Center \\ 
Books I & 2.5M & 20 & 50.0M & aLLMA Lab \\ 
Books II & 131.7M & 4 & 526.8M & LocalDoc \\ 
Blogs & 0.9M & 20 & 17.5M & aLLMA Lab \\ 
Azerbaijani laws & 44M & 6 & 264M & \citep{allma_lab_2024_eqanun} \\ \midrule
Total & 651.1M & - & 2118.2M & - \\ \bottomrule
\end{tabular}
\caption{Data sources used to generate the DOLLMA corpus. English Wikipedia has been translated with open-source models by the BHOS AI team.}
\label{tab:data_sources}
\end{table*}

Due to these limitations, we decided to curate a new dataset specifically for pre-training LLMs that understand Azerbaijani. This new corpus is called DOLLMA (\textbf{D}ataset for \textbf{O}pen \textbf{L}arge \textbf{L}anguage \textbf{M}odels in \textbf{A}zerbaijani).\footnote{\url{https://huggingface.co/datasets/allmalab/DOLLMA}} The first and current version of this dataset contains Azerbaijani Wikipedia, Translated English Wikipedia (incomplete), news, blogs, books, and Azerbaijani laws. This dataset contains about 651.1 million words.\footnote{Words were counted with a simple whitespace tokenizer.} New versions of DOLLMA will incorporate the Common Crawl data.

\textbf{Books.} We attempted to create a large book corpus but faced several challenges. Most of the available electronic books in Azerbaijani are scanned copies. Publishers rarely offer electronic books that are suitable for text extraction. As of 9 May 2024, Qanun Publishing, the largest publishing house in Azerbaijan, offers 52 PDFs or EPUBs on its website. The remaining books, which were sampled from the Azerbaijan National Library \footnote{\url{https://www.millikitabxana.az/}}, Children's Library \footnote{\url{https://www.clb.az/}}, and other sources, are all scanned copies that have occasionally passed through an OCR model. For OCR, Tesseract \citep{Tesseract} was chosen due to its multilingual support and open-source availability. We scanned thousands of books and manually sampled and analyzed them. Tesseract failed to capture guillemets, which is widespread in older Azerbaijani books. It also mixed up "m" with "rn" in scanned books. This happened often enough to decrease the quality of the text substantially. Due to these limitations, we decided against using OCR output altogether as training data. Instead, we opted for two datasets:
\begin{enumerate}
    \item Books I contains a small number of handpicked books.
    \item Books II contains a higher number of books with less detailed processing.
\end{enumerate} 

\textbf{Wikipedia.} We used dumps provided by the Wikimedia Foundation to create a new version of Azerbaijani Wikipedia. Both the data \citep{azwiki} and cleaning scripts \footnote{\url{https://github.com/ceferisbarov/azwiki}} are publicly available. BHOS AI team leads another initiative where they are using open-source translation models to translate English Wikipedia into Azerbaijani \citep{bhos_2024}. While this dataset offers little in terms of linguistic variety, it provides an invaluable knowledge base to train the models. Therefore, it was included in the final corpus.

\textbf{News.} There is an abundance of news datasets for Azerbaijani. However, we decided against using a very large news corpus, since it offers little variety in terms of language.
In our experience, models trained on news datasets do not learn the language comprehensively, possibly because the news contains little to no creative writing, first- and second-person narration, and dialogue. Due to these limitations, only two news datasets were included. One contains text scraped from several news platforms, and the other contains news and updates from Azerbaijan National Library. The BHOS AI team provided both datasets.

\textbf{Blogs.} Another data source was blog posts collected from various websites. Instead of scraping a large number of websites for their blogs, several blogs were manually picked due to their high-quality text and informative content.

\textbf{Laws.} The last part consisted of Azerbaijani laws, all of which are publicly available. We have also released this as an independent text corpus \citep{allma_lab_2024_eqanun}.

You can see a summary of these sources and their accompanying upscaling ratios in Table \ref{tab:data_sources}.
Upscaling ratios were decided rather arbitrarily. We decided against upscaling the news since they offer little linguistic variety. Azerbaijani Wikipedia was upscaled higher than the translated English Wikipedia to account for the lossy translation process. Azerbaijani laws offer higher-quality text than Azerbaijani Wikipedia but offer less variety both in terms of content and form. Considering this, we upscaled them at the same level. Blogs and Books II datasets were hand-picked and constituted the highest-quality text in our corpus. Therefore, their upscaling ratio was the highest. Books II had mediocre quality, mostly due to the challenges of extracting text from PDF files. We upscaled it at the same level as the English Wikipedia.

A major shortcoming of DOLLMA is imbalanced domain distribution. While the dataset contains a substantial amount of text on Azerbaijani laws, it is lacking in terms of first-person narrative, and STEM fields. It is also heavily Azerbaijan-centric, which may or may not be an issue depending on the final goal.

Deduplication has not been performed since none of the sources has the potential of overlapping with another (i.e., Wikipedia and News, or Books and Laws). However, the addition of a deduplication stage is important if this corpus is to be expanded further.

Later versions of DOLLMA will include several major changes:

\begin{enumerate}
    \item Add deduplication to the pipeline. This will allow us to incorporate potentially overlapping text sources.
    \item Create a large-scale book corpus.
    \item Improve domain distribution.
    \item Incorporate web-scraping datasets such as OSCAR and C4.
\end{enumerate}

We believe that these changes will open up new possibilities for modeling the Azerbaijani language. At the current state, however, taking into account time and hardware limitations, our dataset was sufficient to continue to the modeling stage.

\begin{table*}
\centering
\begin{tabular}{lllll}
\toprule
Model & Hidden Size & Num. Attention Heads & Num. Hidden Layers & Num. Parameters \\ \hline
aLLMA-Small & 512 & 8 & 4 & 45.9M  \\
aLLMA-Base & 768 & 12 & 12 & 135.2M  \\ 
aLLMA-Large & 1024 & 16 & 24 & 369.5M \\ \bottomrule
\end{tabular}
\caption{ Architectural differences among the aLLMA models.}
\label{tab:model_sizes}
\end{table*}

\section{Pre-training}

Using DOLLMA, we have developed a series of foundational language models called aLLMA (\textbf{a} \textbf{L}arge \textbf{L}anguage \textbf{M}odel for \textbf{A}zerbaijani). aLLMA has been trained in three sizes: small, base, and large. Base and large correspond to the original BERT models BERT\textsubscript{BASE} and BERT\textsubscript{LARGE} \citep{devlin-etal-2019-bert}. Small architecture was borrowed from \citep{bert-small}. Architectural details of these models can be found in Table \ref{tab:model_sizes}. All three models\footnote{\url{https://huggingface.co/allmalab/bert-small-aze}}$^{,}$\footnote{\url{https://huggingface.co/allmalab/bert-base-aze}}$^{,}$\footnote{\url{https://huggingface.co/allmalab/bert-large-aze}} have been released publicly and included in our benchmarks.

We recognize two alternative approaches to the problem of modeling a low-resource language:
\begin{itemize}
    \item Continue the pertaining step of an existing multilingual foundation model.
    \item Pre-train a foundation model from scratch.
\end{itemize}

aLLMA models were developed with the latter approach. While the benchmarks contain several models that have been trained with the former method, no detailed analysis of the performance difference is provided. This is left as a future research area.

\begin{figure}
    \centering
    \includegraphics[width=\columnwidth]{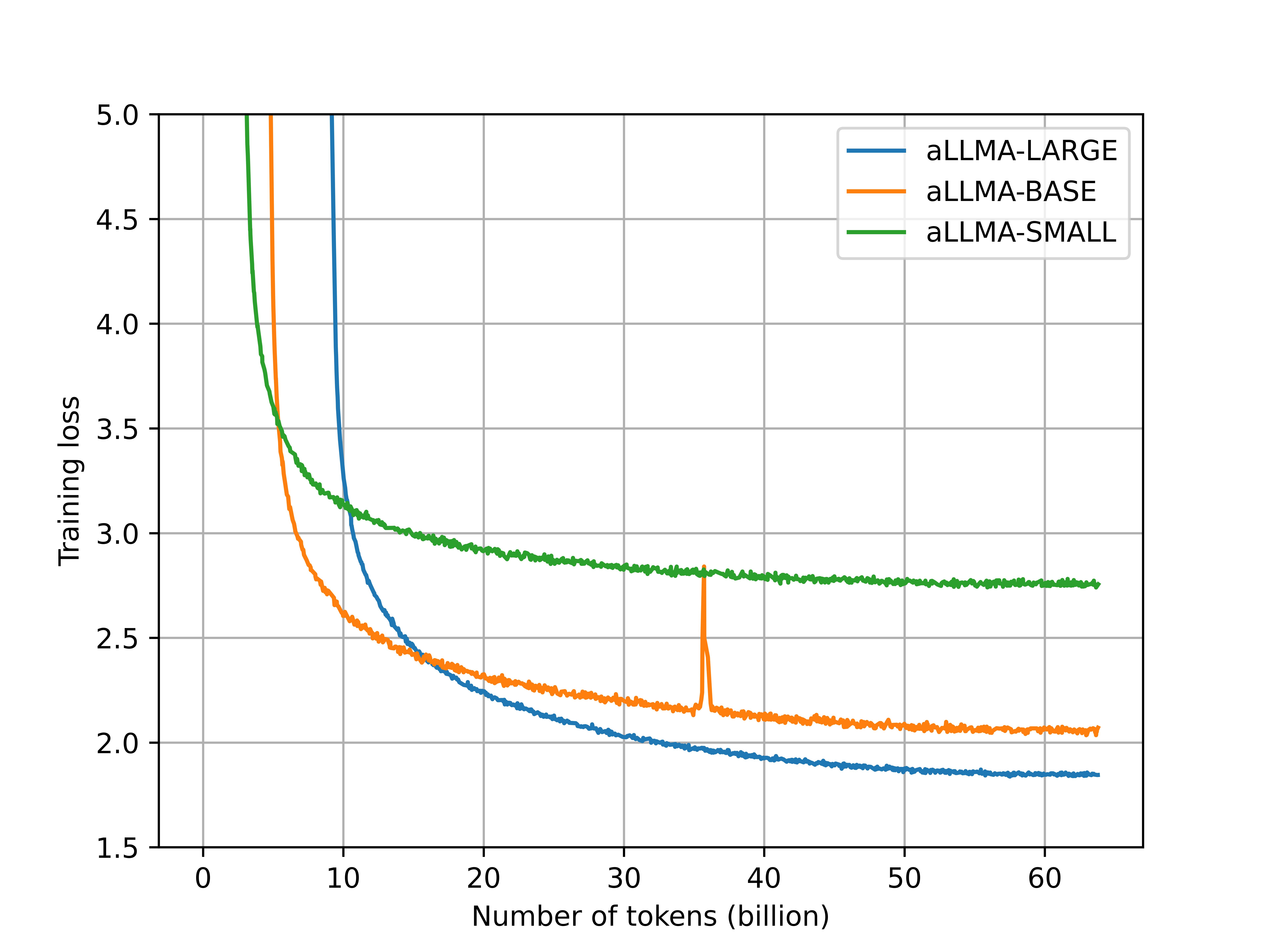}
    \caption{Training loss for aLLMA-Small, aLLMA-Base, and aLLMA-Large models.}
    \label{fig:loss_curve}
\end{figure}

The pre-training task was only masked language modeling. The next sentence prediction task constitutes one of our benchmarks but is not included in the pre-training stage. Training loss of aLLMA-Small and aLLMA-Base models can be found in Figure \ref{fig:loss_curve}. 

One major limitation of the original BERT paper was static masking. If tokens are masked before the training process, then even with multiple epochs, the model will always have to predict the same token. We borrow the idea of dynamic masking from \citep{liu2019roberta}. Instead of masking tokens before the training, tokens are masked on demand. This results in various masking patterns on the same text samples. 
Since our model is trained from scratch on an Azerbaijani-only dataset, using existing multilingual tokenizers offered no advantages. A WordPiece tokenizer\footnote{\url{https://huggingface.co/allmalab/bert-tokenizer-aze}} was trained on a weighted version of DOLLMA, with a vocabulary size of 64k. We have not performed a systematic evaluation to find the optimal vocabulary size. \citep{KAYA2024200335} have researched the impact of vocabulary size on the performance of Turkish language models. Since both Azerbaijani and Turkish are agglutinative languages and share similar morphological features, we used the results of this research as a guide. While \citep{KAYA2024200335} recommends increasing this number further, anything above that would be too computationally expensive for us. 

\begin{table*}
\centering
\begin{tabular}{llll}
\toprule
Dataset & Num. of samples & Task & Source \\ \midrule
AZE-SCI & 5.76k & Text classification & \citep{mammad_hajili_azsci} \\
MRPC (translated) & 3.67k & Paraphrase identification & \citep{eljan_mahammadli_2024} \\
WikiANN & 12k & Named entity recognition &  \citep{wikiann} \\ 
SQuAD (Translated) & 54.1k & Extractive QA & \citep{azsquad} \\ 
LDQuAd & 154k & Extractive QA & \citep{localdoc_2024} \\
AZE-NSP & 9.15k & Next sentence prediction & \citep{allma_lab_2024_nsp} \\ \bottomrule
\end{tabular}
\caption{Benchmarks.}
\label{tab:benchmarks}
\end{table*}

\section{Benchmarks}
This section presents the tasks that were used to evaluate the natural language understanding capabilities of foundation models in Azerbaijani. All of these tasks are a form of classification since the models are encoder-only. We created three new datasets - text classification (AZE-SCI), closed-book multiple-choice questions (CB-MCQ), and next-sentence prediction (AZE-NSP) as a part of this project. Four more datasets (WikiANN, translated MRPC, translated SQuAD, and LDQuAd) were borrowed from the open-source community.

For each task, all models were trained with the same hyperparameters (learning rate, number of epochs, etc.). In almost all cases, models were undertrained - the project had hardware and time constraints and we were trying to get comparative results rather than functioning models. The source code for all experiments is being released, and the reader can generate better-performing models by simply training longer. Benchmarks have been summarized in Table \ref{tab:benchmarks}.

\subsection{AZE-SCI}

AZE-SCI dataset contains titles, topics, and subtopics of dissertations written at Azerbaijani universities and institutes. Subtopics were ignored and only topic labels were used for classification. Being the simplest out of all, this dataset offers a traditional text classification challenge. \citep{mammad_hajili_azsci}

\subsection{AZE-NSP}
The next-sentence prediction task allows us to assess the higher-level understanding capabilities of the models. We were unable to find such a dataset in Azerbaijani and decided to build one ourselves. Several books were compiled and split into paragraphs. A sentence pair was extracted from each paragraph and divided into two parts. The second sentence served as the true label, while randomly sampled sentences from other parts of the same book functioned as distractors. Special care was taken to ensure that there was no overlap between this dataset's source text and the pre-training data. \citep{allma_lab_2024_nsp}

\subsection{CB-MCQ}

The most challenging task given to the models was a closed-book multiple-choice question-answering dataset, collected from various websites. Its content is mostly middle- and high-school topics, but also contains topics like a driver's exam and state service examination. \citep{allma_lab_2024_mcq}

All of the tested models failed to learn this model even at a basic level. Due to this, we have decided against testing all models and including them in the leaderboards. This benchmark remains an open challenge for Azerbaijani language modeling. It has been released publicly on the Hugging Face platform to promote further research.

\subsection{Existing datasets}
Several open-source datasets were sampled as an evaluation criterion. Some of these datasets were discarded due to low quality or small size. In the end, we decided on WikiANN, translated SQuAD, LDQuAd, and translated MRPC.

\subsubsection{WikiANN}

WikiANN is a multilingual named entity recognition dataset sampled from Wikipedia articles \citep{wikiann}. The dataset contains 12 thousand samples in Azerbaijani. The text is tokenized and location, person, and organization entities are labeled. Since the tokenized version of the dataset does not match our tokenizer, each token was re-tokenized separately and a tag was assigned to each new token.

\subsubsection{SQuAD}

Question-answering problems usually demand more robust language understanding and therefore serve as a better criterion than simpler classification tasks. There is no original open-book question-answering dataset in Azerbaijani. The Stanford Question Answering Dataset (SQuAD) is one such dataset in English. We used a translated and reindexed version of the original \citep{azsquad}.

\subsubsection{LDQuAd}
LDQuAd is a native Azerbaijani alternative to the SQuAD dataset. It contains 154,000 thousand samples, about 30\% of which have no answer. Upon further inspection, we realized that most samples with a "no answer" label actually had a correct answer. It is possible that indices were generated automatically with a string search, and some answers were not found, resulting in mislabeled samples. Due to this, we discarded all samples with no answer. \citep{localdoc_2024}

\subsubsection{MRPC}
Microsoft Research Paraphrase Corpus (MRPC) \citep{mrpc} is an English dataset that is used in NLU benchmarks like GLUE. Each sample contains two sentences and a label of whether or not two sentences are paraphrased versions of each other. We used a translated version of the corpus \citep{eljan_mahammadli_2024}.

\section{Results}

\begin{table*}[]

\centering

\begin{tabular}{lllllllll}

\toprule

Model name & Size & AZE-SCI & MRPC & WikiANN & SQuAD & AZE-NSP & LDQuAd & Avg. \\ \midrule

{\color[HTML]{3166FF} XLM-RoBERTa-Large} & 560M & 89.76 & 82.41 & 92.35 & \textbf{75.70} & 33.46 & 83.48 & 76.19\\ 

{\color[HTML]{3166FF} mDeBERTa-v3} & 279M & 87.13 & \textbf{83.71} & 91.87 & 72.27 & \textbf{78.84} & 85.29 & 83.19 \\ 

{\color[HTML]{ff4f00} mDEBERTA-v3-AZE} & 279M & 89.73 & 80.18 & 91.83 & 70.31 & 78.29 & 85.07 & 82.57 \\

{\color[HTML]{3166FF} XLM-RoBERTa-Base} & 278M & 86.99 & 70.90 & 90.29 & 70.97 & 74.96 & 85.17 & 79.88 \\ 

{\color[HTML]{ff4f00} RoBERTa-Base-AZE} & 278M & 89.17 & 81.25 & 91.62 & 70.36 & 76.98 & 85.44 & 82.47 \\ 

{\color[HTML]{ff4f00} BERT-Base-AZE} & 178M & 88.80 & 80.12 & \textbf{92.35} & 69.42 & 74.12 & 64.41 & 78.20 \\ 

{\color[HTML]{3166FF} BERT-Base-Multi}  & 178M & 86.88 & 79.92 & 91.67 & 68.92 & 72.46 & 83.48 & 80.56 \\ 

{\color[HTML]{000000} BERT-Scratch} & 135M & 73.31 & 65.36 & 72.95 & 16.11 & 50.73 & 26.60 & 50.84\\

{\color[HTML]{000000} BERT-Base} & 108M & 76.73 & 75.00 & 90.94 & 55.51 & 62.12 & 74.88 & 72.53 \\ \midrule 

{\color[HTML]{008000} ALLMA-Large} & 370M & \textbf{91.46} & 81.55 & 91.71 & 73.77 & 78.58 & \textbf{85.93} & \textbf{83.83} \\ 

{\color[HTML]{008000} ALLMA-Base} & 135M & 90.84 & 79.74 & 91.26 & 71.30 & 75.95 & 85.69 & 82.46 \\ 

{\color[HTML]{008000} ALLMA-Small} & 46M & 88.06 & 71.77 & 90.07 & 59.89 & 70.23 & 80.80 & 76.80 \\ \bottomrule

\end{tabular}

\caption{Azerbaijani NLU benchmark. All metrics are F1 score. {\color[HTML]{3166FF} Blue models} are multilingual. {\color[HTML]{ff4f00}Orange models} are multilingual models that have been further pre-trained for Azerbaijani. {\color[HTML]{008000}Green models} were trained from scratch only for Azerbaijani. {\color[HTML]{000000} Black models} serve as baseline.}

\label{benhcmark_table}

\end{table*}

Initial tests were performed on dozens of foundation models and some were deliberately left out of the final analysis due to their inferior performance. The final benchmark includes four model categories:

\textbf{Multilingual foundation models.} 
BERT-Base-MULTI is a multilingual version of the original BERT model. XLM-RoBERTa-Base and XLM-RoBERTa-Large are some of the best-performing multilingual models \citep{xlm-roberta}. mDeBERTa-v3 is a multilingual version of DeBERTa v3 model \citep{he2023debertav3}).

\textbf{Multilingual models further pre-trained for Azerbaijani.} BERT-Base-AZE \citep{mammad_hajili_2024_bert}, RoBERTa-Base-AZE \citep{mammad_hajili_roberta}, and mDEBERTA-v3-AZE \citep{mammad_hajili_2024_debertav2} have been further pre-trained on a small and high-quality Azerbaijani dataset. Their base models are RoBERTA-Base, BERT-Base-MULTI, and DeBERTa-Base, respectively.

\textbf{Models pre-trained from scratch.}
aLLMA-Small, aLLMA-Base, and aLLMA-Large are the only monolingual Azerbaijani models.

\textbf{Baseline models.} The original English-only BERT-Base was added as a baseline for the multilingual models. BERT-Scratch refers to the models trained on a specific task without pre-training weights. It functions as a baseline for all models in the benchmark.

\begin{figure}
    \centering
    \includegraphics[width=\columnwidth]{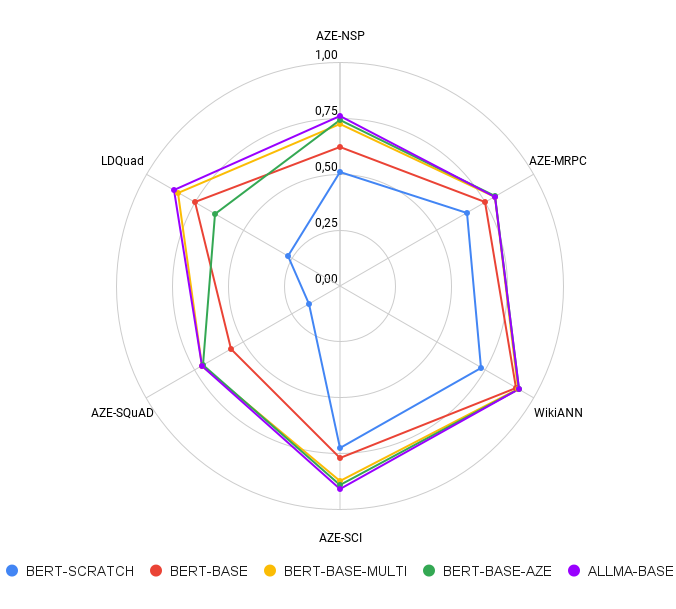}
    \caption{Performance comparison among BERT models of the same configuration. aLLMA-Base outperforms the other models in 4 out of 6 benchmarks.}
    \label{fig:bert_performance}
\end{figure}

You can find the results in Table \ref{benhcmark_table}. mDeBERTa-v3 and aLLMA-Base have the best overall performance. Figure \ref{fig:bert_performance} compares the performance of Base models.\footnote{The difference in number of parameters between these models is due to varying vocabulary sizes. Otherwise, their architectures are identical.} aLLMA-Base outperforms all other models of similar size in 4 out of 6 benchmarks. Comparing BERT-Base-AZE with BERT-Base-MULTI shows that further pre-training of multilingual models can result in some performance improvement, but also model collapse (compare their performance in LDQuAd benchmark). However, a more comprehensive analysis is required before we can make generalizations about the effects of continued monolingual pre-training on multilingual models.

BERT-Scratch performs particularly well on AZE-SCI, MRPC, and WikiANN tasks. We believe this has two explanations. The first is that these tasks can be solved partially with statistical information from the input text, while this is not possible with the other tasks. The second is that the random baseline in these tasks is relatively high, while SQuAD and LDQuAd have very low random baselines.

These results demonstrate several points regarding foundation models for low-resource languages:

\begin{enumerate}
    \item \textit{Pre-training from scratch on a monolingual dataset is a viable strategy for building a low-resource LLM.} aLLMA-Base has competitive performance against larger models despite being trained only on the DOLLMA corpus.
    \item \textit{Multilingual models offer competitive performance even in languages that they were undertrained for.} Azerbaijani has not been the focus in any of these multilingual models (XLM-RoBERTa, mDeBERTa-v3, or BERT-Base-MULTI). Despite this, they outperform most models in some tasks.
    \item \textit{Even monolingual English foundation models can be useful for fine-tuning on a downstream task and perform better than training a model from scratch.} BERT-Base was included in our research as a baseline but exceeded our expectations. This suggests that the state-of-the-art English models can be utilized for certain NLU tasks in Azerbaijani. This remains a potential research area.
\end{enumerate}

It is still possible that we have missed some high-quality models and we are open to feedback regarding this. Our work can be strengthened by finding or creating new benchmarks. We hope that this work will lay the foundations for such developments.

\section{Conclusion}
Despite some academic and community attempts to create a foundation model for Azerbaijani, this problem has not received systemic treatment. We tackle this issue by introducing a new family of foundation models for the language and benchmarking these models and other existing alternatives. To compensate for the lack of datasets suitable for benchmarking LLMs in Azerbaijani, we introduce text classification, closed-book question-answering, and next-sentence prediction datasets.

This work can be extended in several ways. The simplest improvement would be \textbf{training larger models on larger corpora}. Our project does not achieve this due to time and hardware limitations. aLLMA models are not a final product, but an early prototype. A larger training corpus, more advanced hardware, and a better-optimized training process will certainly result in more robust foundation models for Azerbaijani.

A more urgent work, however, is \textbf{extending the benchmarks} by creating more labeled task-specific datasets and adding other existing models to the leaderboards.

\textbf{Including the next-sentence prediction task in the pre-training phase} can increase the performance of aLLMA models further.

Another ambitious direction would be using our corpus to \textbf{develop a generative foundation model.} This paper concentrated on encoder-only models because it is a simpler problem to solve and it has more immediate applications. Nevertheless, generative language models have wide-ranging industrial applications and demand a systemic treatment.

\section*{Acknowledgements}
We thank PRODATA LLC and Microsoft Accelerating Foundation Models Research Program for supporting our research. All of the data and code created as part of this project will be publicly available under permissive licenses.
% \clearpage
\bibliography{anthology,custom}
\bibliographystyle{acl_natbib}

\appendix

% \section{Appendix: Resources}
% \label{sec:appendix}

% datasets
% models
% URLs

\end{document}